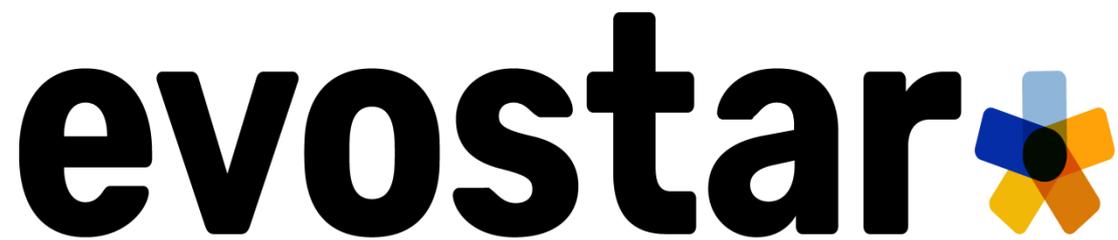

# Evo* 2021

The Leading European Event on

Bio-Inspired Computation

*Online Conference. 7-9 April 2021*

http://www.evostar.org/2021/

## – LATE-BREAKING ABSTRACTS –

**Editors:**

A.M. Mora
A.I. Esparcia-Alcázar

# Preface

This volume contains the Late-Breaking Abstracts accepted at Evo* 2021 Conference, that took place online, from 7 to 9 of April.

They were also presented as short talks as well as at the conference's poster session.

The works present ongoing research and preliminary results investigating on the application of different approaches of Evolutionary Computation and other Nature-Inspired techniques to different problems, most of them real world ones.

These are very promising contributions, since they outline some of the incoming advances and applications in the area of nature-inspired methods, mainly Evolutionary Algorithms.

*Antonio M. Mora*
*Anna I. Esparcia-Alcázar*

# Table of contents

# Parameter control for the Plant Propagation Algorithm[*]

Wouter Vrielink[1][0000−0002−2508−017X]
Daan van den Berg[2][0000−0001−5060−3342]

[1] Informatics Institute
Universiteit van Amsterdam
`WLJ.Vrielink@uva.nl`
[2] Yamasan Science & Education
`kawarimasen0010@gmail.com`

**Abstract.** The plant propagation algorithm, a crossoverless population-based metaheuristic, performs significantly better when its fitness function is deterministically adjusted throughout the optimization process.

**Keywords:** Evolutionary Algorithms · Plant Propagation Algorithm · Metaheuristics · Parameter Control

## 1   PPA & Parameters

Nearly celebrating its tenth anniversary, the plant propagation algorithm (PPA) is still a relative newcomer in the realm of metaheuristic optimization methods. It revolves around the idea that fitter individuals in its population produce many offspring with small mutations, whereas unfitter individuals produce fewer offspring with larger mutations. Since its inception by Abdellah Salhi and Eric Fraga [9], the paradigm has seen a number of applications [14][11][12][15][4][8], as well as some spinoffs [13][7][10][5]. In this paper, we'll use its seminal form, which iterates through the following routine:

1. Initialize *popSize* individuals on the problem's domain with a uniform distribution between the bounds.
2. Normalize each individual's objective value $f(x_i)$ to the interval [0,1] as $z(x_i) = \frac{f(x_{max}) - f(x_i)}{f(x_{max}) - f(x_{min})}$.
3. Assign fitnesses to individuals $x_i$ as $F(x_i) = \frac{1}{2}(tanh(4 \cdot z(x_i) - 2) + 1)$.
4. Assign the number of offspring for each individual $x_i$ as $n(x_i) = \lceil n_{max} F(x_i) r \rceil$, where r is a random value in [0,1) and $n_{max}$ is a parameter determining the number of offspring a population produces.
5. The mutation on dimension $j$ is $(b_j - a_j)d_j(x_i)$, with $d_j(x_i) = 2(r - 0.5)(1 - F(x_i))$, in which $r$ is a random number in [0,1) and $b_j$ and $a_j$ are the upper and lower bounds of the $j^{th}$ dimension. Any individual exceeding a dimension's maximum or minimum bound is corrected back to that bound.
6. The *popSize* best individuals are selected for the next generation. If the predetermined number of evaluations is not met, go back to 2.

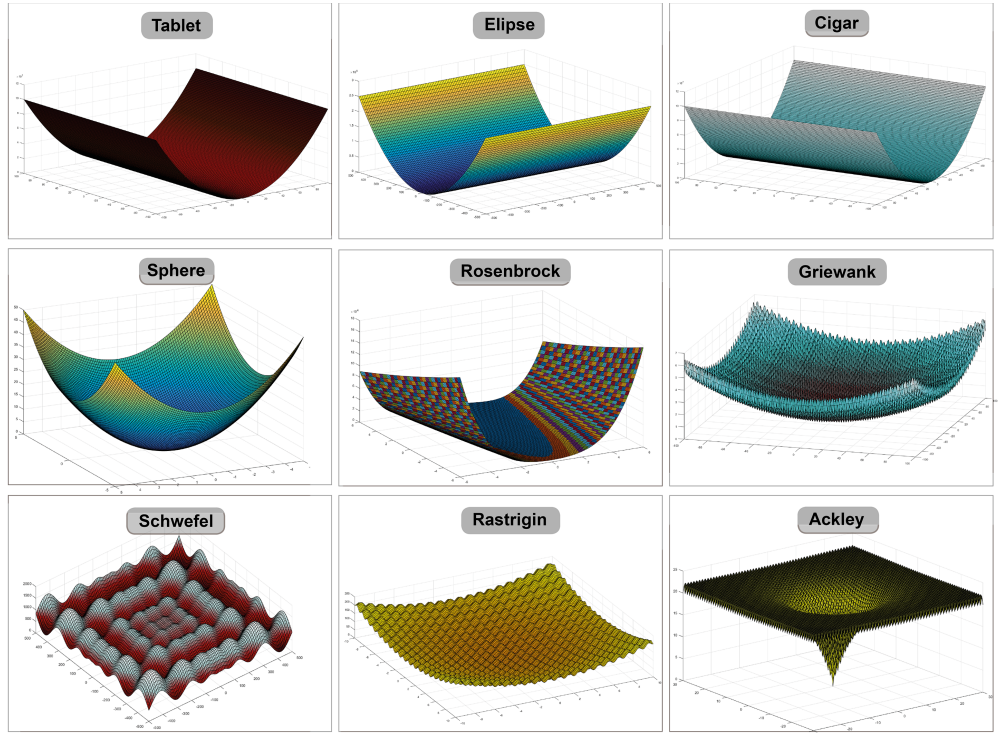

**Fig. 1.** The nine $n$-dimensional benchmark test functions used in this experiment. For all functions in this experiment $n = 2$, and our modified PPA algorithm ran for 10,000 function evaluations.

For this preliminary report, we'll slightly adapt point 3 of the algorithm, the nonlinearization step, which assigns fitness to individuals from the normalized objective values, is adapted to

$$F(x_i) = \frac{1}{2}(tanh(4s \cdot z(x_i) - 2s) + 1) \quad (1)$$

in which $s = \frac{evals}{factor} + 1$. By applying equation this change, $F(x_i)$ starts off as a normally shaped sigmoid, but gradually increases its suddenness, and eventually steepens almost to a step function. Two things are important here. First, $factor$ is a constant $\in \{100, 200, ..., 4000\}$ which determines the slowness of the bending process. Second, $evals$ is the number of completed *function evaluations*, not the number of generations, so the proposed method does not depend on the parameters $popSize$ or $n_{max}$. Even though the algorithm is largely unsensitive to these parameters in isolation [6][2], any interaction with the modification at hand (Eq. 1) is undocumented as yet.

---

[*] This is the first half of a twin submission twin to EvoStar2021, and has an antagonist[16]

In the Eibenistic classification system, a non-stochastic change in parameter during a run such as our modification qualifies as 'deterministic parameter control' [3], but opinions might differ. Its closest of kin could be the cooling schedule in the simulated annealing algorithm, which is shown to exert a great influence on the quality of its end result [1]. In that case, a usually decreasing parameter 'temperature' ascertains the decline in probability that a deteriorative mutation still gets accepted. Like our *s*-parameter, the change of that temperature parameter is typically functionally predetermined through the evaluation number, and would thereby also qualify as 'deterministic parameter control'. The difference of course, is that our parameter control affects the relative fitness within the population, and not some acceptance rate.

## 2   Experiment & Results

We performed a minimization task on nine $n$-dimensional benchmark test functions, with $n = 2$ (Fig. 1). Besides fixating PPA's default parameterizations ($popSize = 30$ and $n_{max} = 5$) throughout the entire experiment, the possible values for $factor \in \{100, 200, 300...4000\}$ accounted for 40 subexperiments on each of the nine benchmark test functions, 360 in total. For each subexperiment, 10 runs of $10^4$ function evaluations was done, and the median end value of a cell's ten runs was recorded (Fig.2).

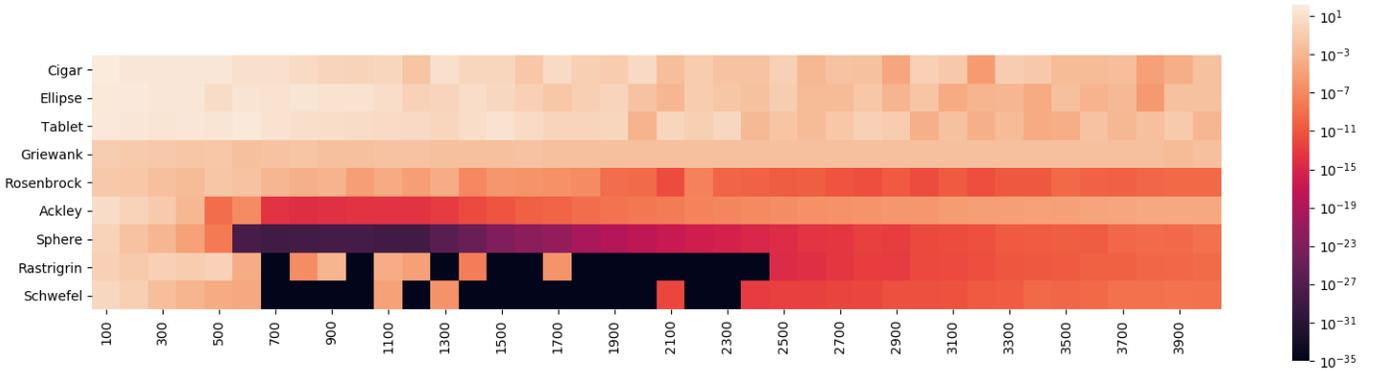

**Fig. 2.** The plant propagation algorithm can improve a lot from incrementing the steepness of its sigmoidal fitness function. The exact improvements seem to depend on *how slow* (horizontally) the fitness function bends, and differs per function.

Three very distinct patterns can be seen. For Cigar, Ellipse, Tablet, Griewank, Rosenbrock and Ackley, our modification results in mild to medium improvements. For the Sphere function, in many ways the simplest in the test suite, there is a gently sloping procession, leading to an optimum somewhere in the region $600 \leq factor \leq 1200$. For Rastrigin and especially Schwefel, there are steep and deep 'black holes', all located in the range $700 \leq factor \leq 2400$, in which the algorithm finds the function's global minimum, greatly outperforming all earlier results [15].

These early results show that it is possible, with negligible computational overhead, to significantly improve PPA's performance by just adding a steepening parameter to its fitness function. How robust and widely applicable these results are however, and if further improvements are possible, still remains to be seen.

# A Dynamic Parameter for the Plant Propagation Algorithm[*]

Wouter Vrielink[1][0000−0002−2508−017X]
Daan van den Berg[2][0000−0001−5060−3342]

[1] Informatics Institute
Universiteit van Amsterdam
WLJ.Vrielink@uva.nl
[2] Yamasan Science & Education
kawarimasen0010@gmail.com

**Abstract.** In the plant propagation algorithm, we deploy an adaptive version of its sigmoidal fitness function which increasingly steepens while running. It increases at a linear rate, and optimal rate windows are identified for the five two-dimensional benchmark test functions, on which the algorithm significantly outperforms earlier results.

**Keywords:** Evolutionary Algorithms · Plant Propagation Algorithm · Metaheuristics · Parameter Control

## 1 PPA & Parameters

At the core, the plant propagation algorithm (PPA) revolves around the idea that fitter individuals in its population produce many offspring with small mutations, whereas unfitter individuals produce fewer offspring with large mutations. Since its creation by Abdellah Salhi and Eric Fraga [9], there have been a number of applications [14][11][12][15][4][8], and adaptations [13][7][10][5], but throughout this, we'll use PPA in its most seminal form:

1. Initialize *popSize* individuals on the problem's domain, uniformly distributed between the bounds.
2. Normalize every individual's objective value $f(x_i)$ to the interval [0,1] as $z(x_i) = \frac{f(x_{max}) - f(x_i)}{f(x_{max}) - f(x_{min})}$.
3. Assign fitness to each individual $x_i$ as $F(x_i) = \frac{1}{2}(tanh(4 \cdot z(x_i) - 2) + 1)$.
4. Assign the number of offspring for each individual $x_i$ as $n(x_i) = \lceil n_{max} F(x_i) r \rceil$, where r is a random number in [0,1), and $n_{max}$ is a parameter determining the number of offspring a population produces.
5. The mutation on dimension $j$ is $(b_j - a_j)d_j(x_i)$, with $d_j(x_i) = 2(r - 0.5)(1 - F(x_i))$, in which $r$ is a random number in [0,1) and $b_j$ and $a_j$ are the upper and lower bounds of the $j^{th}$ dimension. Any individual exceeding a dimension's maximum or minimum bound is corrected back to that bound.
6. The *popSize* best individuals are selected for the next generation. If the predetermined number of evaluations is not met, go back to 2.

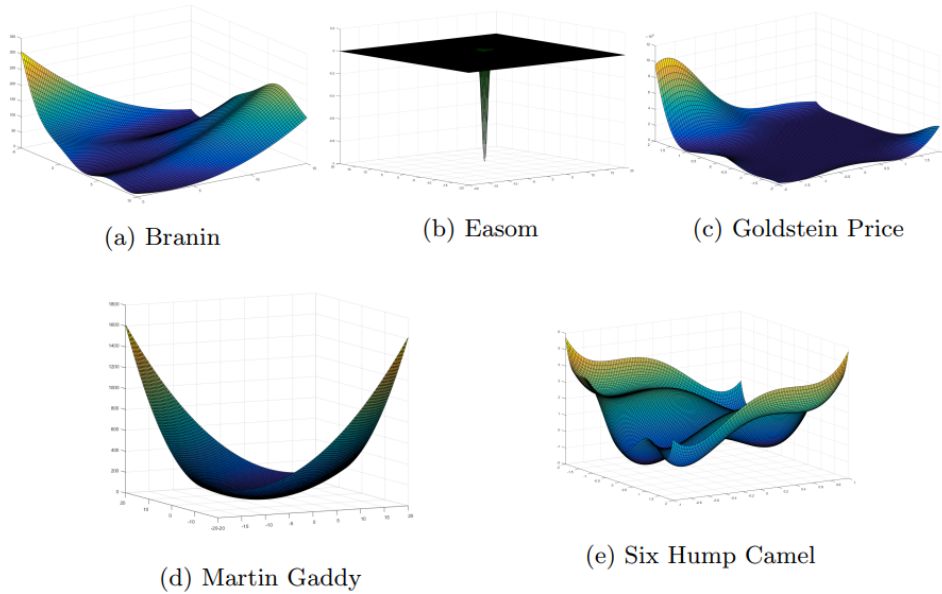

(a) Branin    (b) Easom    (c) Goldstein Price

(d) Martin Gaddy    (e) Six Hump Camel

**Fig. 1.** The five two-dimensional benchmark test functions used in this experiment.

For this preliminary report, we'll slightly adapt point 3 of the algorithm, where the nonlinearization step of assigning the fitness to the individuals from the normalized objective values is adapted to $F(x_i) = \frac{1}{2}(tanh(4s \cdot z(x_i) - 2s) + 1)$ in which $s = \frac{evals}{factor} + 1$. By applying equation this change, $F(x_i)$ starts off as a regular sigmoid, but gently increases its slope until the central derivative is so high the function almost behaves like a step function. Two things are important here. First, $factor$ is a constant $\in \{100, 200...4000\}$ which determines the slowness of the steepening process. Second, $eval$ is the number of completed *function evaluations*, not the number of generations, so the proposed method does not depend on the parameters $popSize$ or $n_{max}$. Even though the algorithm is largely insensitive to these parameters in isolation [6][2], any interaction with the modification at hand is undocumented as yet.

The $s$-parameter reminds a bit of the linear cooling schedule in simulated annealing [1]. In that algorithm, which has a population size of 1, a (usually decreasing) parameter 'temperature' lowers the probability that a deteriorative mutation still gets accepted during an iteration. Like our $s$-parameter, the trajectory of simulated annealing's temperature parameter is usually predetermined before starting a run, and depends only on the number of function evaluations. In the Eibenistic classification system, such non-stochastic change in a parameter during a run qualifies as 'deterministic parameter control' [3].

---

* This is the second half of a twin submission twin to EvoStar2021, and has an antagonist[16]

## 2   Experiment & Results

We performed a minimization task on five two-dimensional benchmark test functions (Fig. 1). Besides fixating PPA's parameters to defaults ($popSize = 30$ and $n_{max} = 5$) throughout the entire experiment, the possible values for $factor \in \{100, 200, 300, ..., 4000\}$ accounted for 40 subexperiments on each of the five benchmark test functions. For each of these 200 subexperiments, 10 runs of $10^4$ function evaluations were done, and the median end value of each subexperiment's ten runs was taken to the corresponding cell in the heatmap of Fig.2.

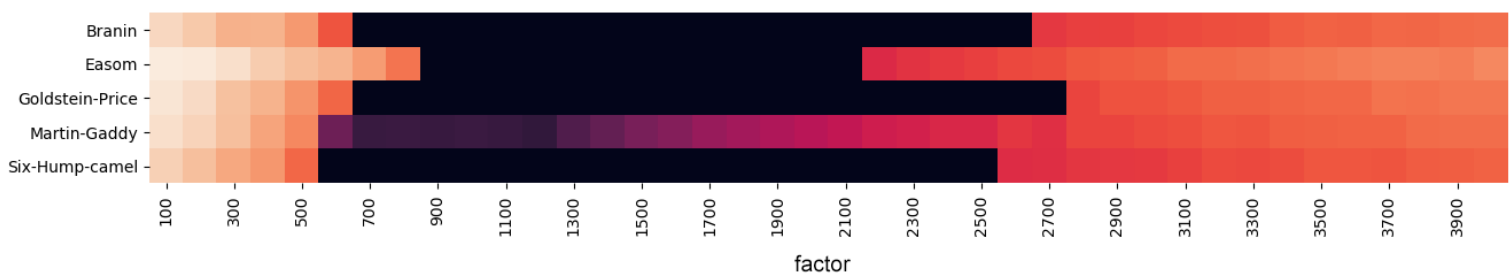

**Fig. 2.** The Plant Propagation Algorithm can improve a lot from deterministic parameter control. The exact improvements seem to depend on *how fast* the fitness function bends, and the effect differs per benchmark test function, but an 'optimal window' is clearly visible. The darker a cell, the closer the median of its ten runs is to the global minimum. The color scale is logarithmic, emphasizing the sharp edges of the black window.

Two distinct patterns can be seen. For Easom, Six-Hump Camel, Branin and Goldstein-Price, the optimal parameter value manifests itself as a deep black crevice roughly located in the range $600 \leq factor \leq 2700$. Only For the Martin-Gaddy function, there is a more gently sloping procession, leading to an optimum somewhere in the region $400 \leq factor \leq 1200$. These early results show that it is possible, at least for these 2D benchmark test functions functions, to significantly improve PPA's performance, with hardly any additional computation, just by adding a steepening parameter to its fitness function. How robust and widely applicable these results are, and if further improvements are possible, still remains to be seen.

Plant propagation is known as a relatively stoic algorithm. Its minimizing capabilities are relatively constant across a large set of benchmark test functions [15]. It is also largely insensitive to parameter variations in $popSize$ and $n_{max}$ on these five 2D-benchmark functions [6], as well as nine other n-dimensionals [2]. However, in the latter, the sensitivity development of the algorithm across dimensionalities differed from function to function. The authors of that study conjectured that these difference might originate from the constituents of the functions themselves, being exponentials, quadratics, or trigonometrics.

This does not appear to be the case for the five functions from this study. All aforementioned constituents are present in these five benchmark test functions, and their approximability for deterministically parameter controlled PPA can not be separated on that basis. The destinction can still be sought in functional characteristics, but may be found in a different property, or a different domain.

# Sensitivity to Partial Lamarckism in a Memetic Algorithm for Constrained Portfolio Optimization*

Feijoo Colomine Durán[1], Carlos Cotta[2,3], and Antonio J. Fernández-Leiva[2,3]

[1] Universidad Nacional Experimental del Táchira (UNET), Laboratorio de Computación de Alto Rendimiento (LCAR), San Cristóbal, Venezuela
`fcolomin@unet.edu.ve`
[2] Dept. Lenguajes y Ciencias de la Computación, ETSI Informática, University of Málaga, Campus de Teatinos, 29071 - Málaga, Spain {`ccottap,afdez`}`@lcc.uma.es`
[3] ITIS Software, Universidad de Málaga, Spain

**Abstract.** We analyze the sensitivity of a memetic algorithm to the parameter governing the stochastic application of local search (based on Sharpe index) in the context of constrained portfolio optimization, and compare it to non-memetic proposals.

**Keywords:** memetic algorithms, portfolio optimization, Sharpe Index, multi-objective optimization.

## 1 Introduction

The problem of portfolio selection amounts to determining the structure of an investment portfolio, that is, given a number of financial instruments deciding the share of each of them in the portfolio. This is typically done based on the historical evolution of assets traded on the stock market and how these are influenced by their own management capacity or affected by the environment in which they operate. Markowitz model is a cornerstone in this context, providing a framework to characterize the performance an risk of an investment portfolio. Since its application results in a non-linear parametric problem that is very difficult to solve using analytical techniques, heuristic optimization methods are particularly appropriate in this scenario. In this sense, we proposed several evolutionary methods to tackle this problem, building on Sharpe index [6] as a way to measure the risk-performance ratio of investment portfolios [3]. We focus now on augmenting these evolutionary methods with a local search component, hence resulting in a memetic algorithm. An issue of interest is the effect of this component and what the sensibility of the algorithm is to its rate of application. This work focuses on this, analyzing the performance of the MA is with respect to this parameter, and how it compares to non-memetic counterparts.

---

* This work is partially supported by MinEco under project TIN2017-85727-C4-1-P. C.C. acknowledges Australian Research Council's Discovery Project DP200102364.

## 2   Materials and Methods

We consider the use of multi-objective evolutionary algorithms for optimizing portfolios under cardinality constraints in the context of Markowitz model. We have considered a memetic approach by adding a knowledge-based decision-supported local search to IBEA, a well-known multiobjective evolutionary algorithm that performs satisfactorily on this problem [3]. The application of this intensification component is controlled by a parameter $P_{LS}$ that determines whether it is activated on any newly generated solution –see algorithm 1– hence resulting in partial Lamarckism [5] whenever $0 < P_{LS} < 1$. Solutions are defined by the enumeration of the weights of each fund in the portfolio. These weights are encoded as binary strings and local search is conducted on these solutions by performing first-ascent hill climbing (using Sharpe index as the acceptance criterion). This operator is applied for $\max_{\mathcal{N}}$ iterations.

---

**Algorithm 1:** Sharpe-based Memetic Algorithm

**Parameter:** $\mu$: population size; $\lambda$: # of offspring; $p_{LS}$: local-search probability

```
1  for i ← 1 to μ do
2  |   pop_i ← GenerateSolution();
3  end for
4  while ¬ Termination() do
5  |   offspring ← SelectAndReproduce(pop);
6  |   for i ← 1 to λ do
7  |   |   if rand < p_LS then
8  |   |   |   offspring_i ← LocalSearch(offspring_i);
9  |   end for
10 |   pop ← Replace (pop, offspring);
11 end while
```

---

## 3   Experimentation

The data used in the experiments corresponds to 20 securities from the Colombian Stock Exchange [2], sampled in the period between 2010 and 2016. Portfolios are constrained to have $K = 14$ funds. Experiments have been carried out with the MA by varying the probability of activation of the local search $P_{LS}$ in order to test the sensibility to this parameter. More precisely, we have considered values $P_{LS} \in \{0, 0.05, 0.10, 0.20, 0.30, 0.50, 0.80, 0.90, 1.00\}$. Note that the case $P_{LS} = 0$ corresponds to IBEA. For comparison purposes, we have also included in the experimentation NSGA-II [4] and SPEA2 [7]. We have used the PISA library [1] for basic multi-objective EAs, as well as for the evolutionary engine of MA. In all cases, weights are encoded using 10 bits (resulting in bitstrings of length $\ell = 200$ since there are 20 funds), the crossover rate is $P_x = 0.8$, the mutation rate is $P_m = 1/\ell$, and the population size is $2\ell$. Regarding the MA, $\max_\mathcal{N} = 10$. The algorithms have been run 30 times for 20000 fitness evaluations (except the MA, which is reduced to 18500 to account for the overhead of local search). The results are evaluated from a purely multiobjective perspective (taking the final Pareto front of each run and using the hypervolume indicator and the distance to the best-known front obtained by aggregating all results) and from the financial performance point of view (as measured by Sharpe index).

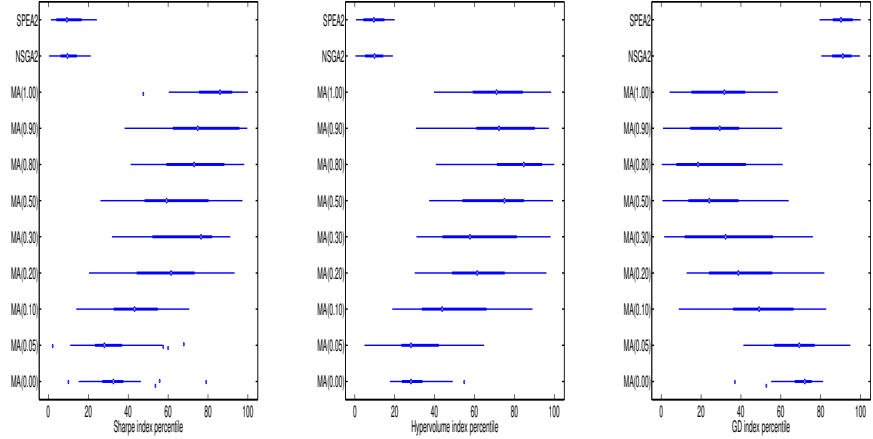

**Fig. 1.** Percentile distribution attained across 30 runs of the different algorithms considered: Sharpe index (left), hypervolume (middle), generational distance (right).

The results are presented in Figure 1 from a comparative standpoint (by considering the aggregation of results from all algorithms and displaying the range of percentiles in which each algorithm performs regarding each indicator; thus, the closer to 100, the better in case of maximization –i.e., Sharpe index and hypervolume– and the closer to 0 the better in case of minimization – i.e., generational distance (GD)). Notice how the performance of the MA is rather correlated with the amount of local search performed. While this is to some extent expected in the case of financial performance (the best results for Sharpe index are obtained for $P_{LS} = 1.0$, and are better –with statistical significance at $\alpha = 0.05$ according to a Wilcoxon ranksum test– to the results of $P_{LS} \leqslant 0.8$), it must be noted that $P_{LS} = 0.8$ provides the best results in a purely multiobjective sense (statistical significance at $\alpha = 0.05$ in all cases except $P_{LS} = 0.9$ for hypervolume, and for $P_{LS} \leqslant 0.3$ for distance). Clearly, the progression in Sharpe index carries along the other measures as well, but in this particular scenario there is an intensification level beyond which the search becomes dominated by this measure, affecting the diversity along the Pareto front. It should also be noted that in all cases the MA outperforms the non-memetic counterparts with statistical significance in all cases.

## 4 Conclusions

A memetic proposal for constrained portfolio optimization has been developed based on the aggregation of a local search whose decision-making is produced through the analysis of the Sharpe index. A sensitivity analysis has been carried out for different levels of performance of the local search described above through probability values coupled with a performance analysis comparing the response with three well-known state of the art in the area, namely: the IBEA, the NSGA2 and the SPEA2. Our results indicate a clear advantage in the application of the memetic strategy in the context of the hypervolume and GD quality indicators and in the performance of the portfolio measured through the Sharpe index.

# Quantum fitness sharing in memetic algorithms for level design in Metroidvania games[⋆]

Álvaro Gutiérrez Rodríguez[1], C. Cotta[1,2], and Antonio J. Fernández-Leiva[1,2]

[1] Dept. Lenguajes y Ciencias de la Computación, ETSI Informática, Universidad de Málaga, Campus de Teatinos, 29071 - Málaga, Spain {ccottap,afdez}@lcc.uma.es
[2] ITIS Software, Universidad de Málaga, Spain

**Abstract.** This work presents an improvement of a procedural content generation system with general purposes but adapted to the design of levels in Metroidvania games using a model of the preferences and experience of the designers. This system obtains promising results about the accomplishment of the designer preferences, but the generated diversity is lower. Therefore, we propose the addition of the fitness sharing and a local search process to obtain quality and diverse solutions. In the calculation of the sharing value we need a distance value between individuals, therefore, we propose the use of a quantum circuit to get this distance.

**Keywords:** Quantum Programming, Evolutionary Algorithm, Memetic Algorithm, Procedural Content Generation

## 1 Introduction

The monotony in videogames is a problem that could make players who finished all levels in a game lose interest in replaying it. As consequence, videogame studios need to create a diverse gaming experience that generates in the players the need to replay the complete video game. These studios use Procedural Content Generation (PCG) techniques to break with this monotony and to assist in the design and development phases. Real applications of these techniques ranges from procedural generation of maps[3] or weapons[4] to stories[5]. The generated content by these techniques is typically well-received by the end user.

AI-Assisted tools for videogame design have been developed with very promising results [1,3,4,5,6,8], but most of these works have focused in a certain genre or in a certain design phase. In this work, we extend a general AI-Assisted tool

---

[⋆] This work is supported by Spanish Ministerio de Economía, Industria y Competitividad under project DeepBIO (TIN2017-85727-C4-1-P), and by Universidad de Málaga, Campus de Excelencia Internacional Andalucía Tech. C.C. acknowledges the Australian Research Council's Discovery Project DP200102364.

[3] http://spelunkyworld.com/index.html
[4] https://borderlandsthegame.com/
[5] https://www.shadowofwar.com/es/

that can help in any design phase of any videogame genre [2]. This proposal uses a PCG technique which uses two components: a learning component and an optimization component. The learning component is an artificial neural network (ANN) that learns designs labeled as *good* or *bad* by the designer, and the optimization component is an evolutionary algorithm that uses the learning component in the assessment process. The final objective is to generate new content based on the previous learning.

Although good results were obtained with this approach, the content generated by the system did not have a good level of diversity. For this reason, we propose the improvement of the system by augmenting the optimization component with a local search process and a new fitness sharing as assessment function. With these improvements, we obtain a memetic algorithm [7] which we expect to find more than one local optimum.

Since the fitness sharing needs a distance calculation measure to penalize closer individuals, we propose the use of a quantum circuit developed in [9]. The rationale for this proposal is the fact that the Euclidean distance (the most commonly used choice for this purpose) loses precision when working with vectors of large dimensionality. With this replacement, we are moving from the Euclidean space to the Hilbert space, that is, we can build the same information with more detail. This will help us to study the proximity of the individuals.

## 2  Local search process

The local search process evaluates the closest neighbors (*Hamming* distance of 1) of an individual and keeps the one that has the best quality value. The so-defined neighbors are incrementally evaluated by focusing on the modified solution parts, reducing computation time. Once the best neighbor is selected, the process would be applied until no neighbor provides an improvement or the process has been applied a maximum number of times.

## 3  Fitness sharing

The fitness sharing is defined as

$$f_{sharing}(i) = \frac{f_{ANN}(i)}{\sum_{j=1}^{\mu} sh(qd_{ij})} \quad (1)$$

where $sh(qd_{ij})$ is the *sharing* value of the quantum distance of the individuals $i$ and $j$, $\mu$ is the population size and the $f_{ANN}(i)$ is the *raw* value of the individual $i$ returned by an ANN.

The *sharing* value, $sh$, is defined as

$$sh(d_{ij}) = \begin{cases} 1 - \left(\frac{qd_{ij}}{\sigma_{sharing}}\right)^{\alpha} & \text{if } d_{ij} < \sigma_{sharing}, \\ 0 & \text{otherwise,} \end{cases} \quad (2)$$

where $\sigma_{sharing}$ is the *sharing* radius, $\alpha$ is a weight and $qd_{ij}$ is the quantum distance returned by the quantum circuit.

This function helps to locate peaks on the search space, by forcing the population to spread away from crowded regions (i.e., those with many neighbors within the sharing radius).

## 4 Quantum circuit as a distance calculator

The quantum circuit is a so-called controlled swap (c-swap) or Fredkin ciruit where we have three qubits: an ancilla qubit and the inputs x and y. The inputs of the circuit are the individuals that we want to compare and their information will be encoded by two qubits. When the value of the ancilla qubit is 1, the values of the controlled qubits are swapped, only if their values are different. The implementation, showed in Fig. 1 applies a superposition state to the ancilla qubit with a Hadamard gate, a c-swap entaglement with the inputs and a final Hadamard gate in the ancilla qubit. The information of the input individuals are encoded by their qubits with U3 gates.

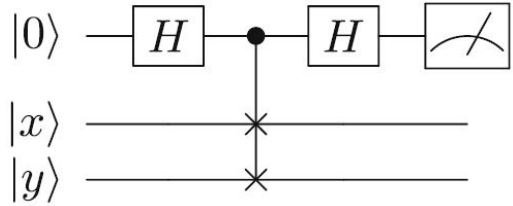

**Fig. 1.** Quantum circuit to the distance calculation

As preliminary experiment, we applied the Euclidean (e) and quantum (q) distance to the following vectors:

| Vector | Label |
|---|---|
| [7, 4, 3, 5, 7, 4, 7, 5, 7, 6] | V1 |
| [4, 5, 2, 6, 3, 6, 5, 2, 0, 0] | V2 |
| [6, 6, 3, 7, 3, 7, 2, 7, 5, 0] | V3 |
| [1, 7, 5, 6, 4, 3, 5, 4, 0, 2] | V4 |

| Vector | V1 | V2 | V3 | V4 |
|---|---|---|---|---|
| V1 | 0.0 | e=11.4 q=0.51 | e=10.14 q=0.47 | e=11.4 q=0.08 |
| V2 | e=11.4 q=0.51 | 0.0 | e=8.18 q=0.02 | e=6.32 q=0.43 |
| V3 | e=10.14 q=0.47 | e=8.18 q=0.02 | 0.0 | e=9.74 q=0.48 |
| V4 | e=11.4 q=0.08 | e=6.32 q=0.43 | e=9.74 q=0.48 | 0.0 |

**Table 1.** Results for the preliminary experiment

## 5 Outlook and Future Work

An improvement of an AI-Assisted tool for videogame design is presented. This improvement has the objective to get more diversity of the final individuals and the populations generated by the optimization component. To this end, a local search process and a fitness sharing have been added to the optimization process. The calculation of the fitness sharing needs a similarity metric and we propose to use a quantum circuit as the calculation of this metric.

Once the study of the theoretical viability of this improvement has been conducted, the new version of the system must be implemented and exposed in a real environment to confirm the initial hypothesis of adding more diversity in the final individuals and the populations generated by the optimization component.

# CLIP-Guided GAN Image Generation: An Artistic Exploration

Amy Smith[1] and Simon Colton[1,2]

[1] Game AI Group, EECS, Queen Mary University of London, UK
[2] SensiLab, Monash University, Australia

## 1 Introduction

The task of text-to-image generation [1] requires the automatic generation of images which reflect a given word or phrase, normally in terms of content, but also potentially in terms of other visual aspects. Generative adversarial networks (GANs) have been shown to be effective for the generation of images containing certain content [5]. As an example, the BigGAN model [3] is able to generate realistic images with content from one of a thousand categories, given a random latent vector input and an appropriate one-hot class vector.

In early 2021, a new approach to finding latent vectors for GANs was developed by members of an online tech/art community sharing their code through Colab notebooks [2]. The approach employs the CLIP pair of pre-trained models [4], which both calculate a latent vector encoding for a given input. For the first CLIP model, the input is a piece of text, and for the second, it is an image. CLIP has been trained so that the latent encoding of pairs (image, text) where the text appropriately describes the image – or vice versa, where the image is an appropriate response to the text – have a lower cosine distance than pairs where the match is not appropriate. CLIP was trained on 400 million (text, image) pairs scraped from the internet, and has captured a broad and deep understanding of the correlation between text and images, covering visual elements such as content, mood, texture, pattern, lighting, emotion and genre, as well as individual objects or people (if there are numerous examples on the internet).

In the context of text-to-image generation, for a given text prompt, $T$, CLIP can calculate a score for the appropriateness of a generated image, $I$, in terms of the cosine distance between the latent encodings CLIP produces for $T$ and $I$. Hence, a search can be undertaken to find a latent vector input to a GAN which produces an image $I$ with a CLIP encoding as close as possible to the encoding of $T$. Note that it is also possible to search for a generated image to match a given image or some combination of text and image. Various search techniques have been tried in the community, with seemingly the most successful being to use backpropagation, i.e., to start with a random latent vector and training it by propagating the value of a loss function based on CLIP. Such an approach was implemented into a Colab notebook called The Big Sleep by Ryan Murdock (see below), which uses CLIP to search for latent vector inputs to BigGAN. We have found this technique remarkably good for a range of text prompts, ranging from specific content to vague moods, as highlighted in figure 1. We see that not only can the process produce appropriate, artistic images for a text prompt and a text+image pair as prompt, it can also innovate, e.g., some of the musicians heads in the first image of figure 1 are musical notes.

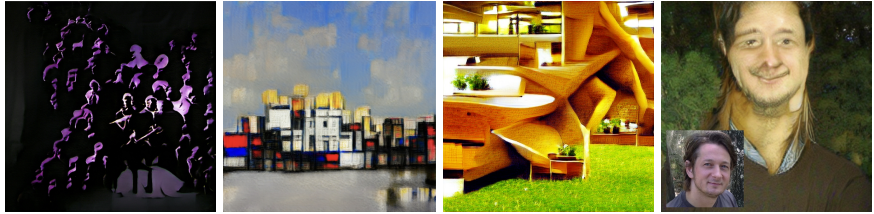

**Fig. 1.** Results from CLIP-guided BigGAN. First three, text prompts only: "Musicians in purple silhouette", "A skyline in the style of Mondrian" and "Architecture is really about well-being". Fourth: guide image inset and text prompt: "Like a horse".

We describe here the usage of four different CLIP-guided GAN image generators via Colab notebooks, to highlight their potential for artistic practice. The first author is trained in the fine arts to degree level and also has a longstanding commercial art practice. We describe here a project where the first author used a combination of the Colab notebooks to produce a pool of themed artistic images. The results were promising in that the artist was largely satisfied with both the results and the process. We compare and contrast the output and processing of the notebooks, and reflect on their potential for art practice in general.

## 2   Production of a Series of Artistic Images

Four CLIP-guided generative models were used 12 times each, resulting in a batch of 48 images. The notebooks were: BigGAN ('The Big Sleep'),[1] DALL-E ('Aleph'),[2] Siren[3] and Lucent's FFT ('Aphantasia').[4] The text prompts used were chosen to be neither too abstract, e.g., "An eyeball in love", nor too mundane, e.g., "an apple on a table", so more artistic and interpretable results could be produced. The prompts used were designed to probe the models' capacity to produce images appropriate to the theme of the series, which was chosen as 'holographic'. The prompts used were: 1) "A painting of holographic gravity", 2) "A painting of a pastel holographic skull", 3) "A holographic crucifix", and 4) "A painting of a holographic dragon". Three images were produced per prompt, per model. We found that, of the 48 images, just 3 were completely unusable, and 39 included content that obviously related to the prompts. This constituted a pool of themed artistic images, some of which are shown in table 1. From this pool, a four part series for potential exhibition was curated, entitled: 'The Holograph Series', as shown in figure 2.

---

[1] colab.research.google.com/drive/1NCceX2mbiKOSlAd_o7IU7nA9UskKN5WR
[2] colab.research.google.com/drive/1oA1fZP7N1uPBxwbGIvOEXbTsq2ORa9vb
[3] colab.research.google.com/drive/1L14q4To5rMK8q2E6whOibQBnPnVbRJ_7
[4] colab.research.google.com/github/eps696/aphantasia/blob/master/Aphantasia.ipynb

| Prompt | Gravity | Skull | Crucifix | Dragon |
|---|---|---|---|---|
| CLIP+DALL-E | 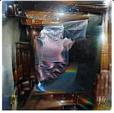 | 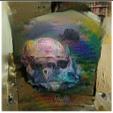 | 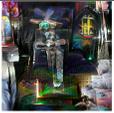 | 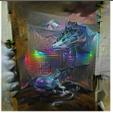 |
| CLIP+FFT | 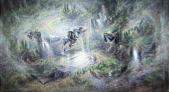 | 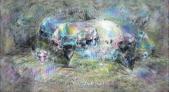 | 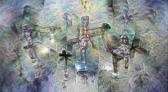 | 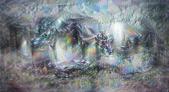 |
| CLIP+BigGAN | 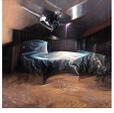 | 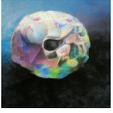 | 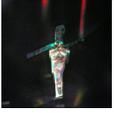 | 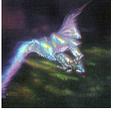 |
| CLIP+Siren | 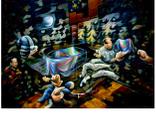 | 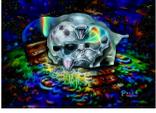 | 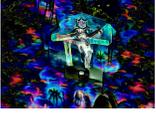 | 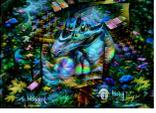 |

**Table 1.** Output images from four CLIP-guided GAN models with the same four holograph-themed text prompts.

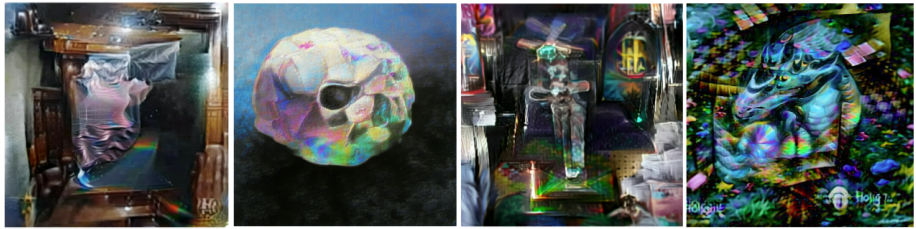

**Fig. 2.** The *Holograph Series* made with CLIP-guided GAN image generators.

Four images specifically were chosen for the series, as each image represents a different prompt used. The images correlate from left to right with the prompts in the order that they are listed above. These final images for the series were chosen based on how well each captured the theme 'holographic' in combination with the subject matter given in each prompt, and how well they fitted visually with one another. Note that no image from the FFT model was used in the series due to the aspect ratio of the output differing substantially to the output from the other models, and the dissonance that occurred between the visual styles of each model, as the FFT images were very noisy and monochromatic. It is clear that all of the models are able to innovate visual content based on the given prompts, which is important when considering the artistic value of the outputs. There is also an impressive degree of diversity in how each model interprets the prompts. This is crucial for producing novel combinations of visual ideas, as the diversity opens up the possibility space, broadening the scope for visual solutions for depicting the contents of the prompts.

For comparison, in table 1, one image out of the three produced for each prompt is presented to examine the diversity of the output from each model, and also to contrast the differing visual traits produced by each. The CLIP+FFT model yielded the least visually diverse results in terms of colour, shape and texture, whilst CLIP+BigGAN (BigSleep) produced very visually diverse results that nevertheless adhered well to the specifications of the prompt. CLIP+DALL-E was close behind, with arguably the most interesting use of colour. Both CLIP+SIREN and CLIP+FFT produced very distinct visual styles that show consistently throughout the results. CLIP+FFT produces uniform centralised compositions that are highly textured, kinetic and de-saturated. CLIP+SIREN contrasts this with high levels of saturation, varying amounts of background texture, and more fully formed compositional elements (namely a dragon head, skull shapes including sockets, and crosses).

## 3   Conclusions and Future Work

We have found that CLIP-guided GAN text-to-image generation produces images which are routinely appropriate to the prompt, usually have artistic value, are often innovative internally and have high diversity across a population for the same prompt. Moreover, the Colab notebooks are highly accessible as they require no technical background, and hence could rapidly influence mainstream art production, given the evidence of valuable usage presented here. One frustration about two of the notebooks is the lack of a stopping condition, which leads to a fear of missing out and extended sessions. We are currently working on more advanced stopping conditions based on thresholds and plateau detection with respect to training loss. Our ultimate aim is to develop a system that feels like a Google image search, but with generated rather than retrieved images.

## Acknowledgements

Many thanks to the Colab notebook authors: Ryan Murdock and Vadim Epstein, for making this technology successful, accessible and exciting. Amy Smith is supported by the UKRI IGGI centre for doctoral training [EP/S022325].

# Short-term effects of weight initialization functions in Deep NeuroEvolution[*]

Lucas Gabriel Coimbra Evangelista[1] and Rafael Giusti[1]

Federal University of Amazonas, Computing Institute, Manaus, Brazil
http://icomp.ufam.edu.br
{lucas.evangelista,rgiusti}@icomp.ufam.edu.br

**Abstract.** Evolutionary computation has risen as a promising approach to propose neural network architectures without human interference, relieving researchers and practitioners from considerable efforts related to the many possible options of parameters and hyper-parameters. However, the high computational cost from that approach is a serious challenge for the application and even research of such methods. In this work, we address this issue by empirically analyzing the short-term effects of weight initialization strategies on the performance of neural networks that creates a highly changing space due to constant interventions from the evolutionary algorithms. We performed experiments with the CoDeepNEAT algorithm on the CIFAR-10 and MNIST datasets. While results are somewhat below start-of-the-art performances, we achieved higher than 99.20% in the MNIST dataset and 84.60% in the CIFAR-10 dataset with significantly less computational time, which suggests that focusing on short-term effects of the weight initialization functions is a promising direction for co-evolution of neural network architectures.

**Keywords:** Evolutionary Computation · Neural Networks · Deep NeuroEvolution.

## 1 Introduction

The Machine Learning community has come a long way since the first deep learning models were proposed. However, in spite of its importance, employing deep models brings two challenges. First, designing neural network architectures is laborious and prone to error, so researchers often restrict themselves to well-known architectures that showed promise in other applications. Second, training these models requires enormous amount of data and computational power.

Weight initialization is an important consideration when designing neural networks, especially with deep models. If done wrong, it might lead to problems, such as vanishing or exploding gradient [1, 2]. Vanishing gradient happens when too small values are chosen by a certain initialization function for the parameters of a neural network, while the exploding gradient problem occurs when the values are too large. Both problems tend to obstruct the learning process, either because the model will converge to a sub-optimal configuration or because the cost function will oscillate around its minimal value instead of converging.

---

[*] This study was financed in part by the Coordenação de Aperfeiçoamento de Pessoal de Nível Superior – Brasil (CAPES) – Finance Code 001

The past decade has been full of important studies seeking to address this problem. Glorot and Bengio [3] found that the logistic sigmoid activation is unsuited for deep networks with random initialization because of its mean value, which can drive especially the top hidden layer into saturation. In turn, they proposed adopting a properly scaled uniform distribution for initialization, the Xavier initialization. Five years later, He et al. [4] found that the Xavier initialization were not suited for deeper models–at the time, at least 8 layers–and proposed a new initialization function, the He initialization.

### 1.1   Deep NeuroEvolution algorithms

Designing neural network architectures requires considerable effort due to the huge possibilities of parameters and hyper-parameters, such as types of layers, learning rate, number of epochs, batch size, variety of initialization functions, whether to use skip connections or not, etc. Considering the advances in the last decade in developing Deep Learning models to fit challenging tasks and the remarkable effort to hand-craft these models, the past years have shown important for advances in methods to propose deep neural network architectures without human interference. These advances were possible from the merge of deep learning field discoveries and bio-inspired algorithms, creating a new area of study: Evolutionary Deep Neural Networks, also called Deep NeuroEvolution.

Papavasileiou et al. [5] calls attention to a novel approach that combines non-gradient descent based evolutionary algorithms and gradient-descent based ones, which could be considered the foundation for the coevolution of Deep NeuroEvolution of Augmenting Topologies (CoDeepNEAT) [6]. In CoDeepNEAT, chunks of layers are evolved as modules and blueprints, which are merged to create multiple architectures. The test results of these architectures are used as fitness values, which are in turn used by CoDeepNEAT to select the best individuals and promote the evolution of new architectures.

## 2   Short-term effect

One important disadvantage of deep learning bio-inspired models is that this approach is very time consuming. In spite of its very competitive results, the time required for evolving the architectures may hinder reproducibility and is a challenge for researchers without access to high computational power, as pointed out by Assunção et al. [7], who used a low-cost GPU for their experiments.

Our motivation comes from an apparent lack of studies that considers the search space of the evolutionary algorithm to be extremely dynamic. In this sense, how do current standard weight initialization functions help convergence of an architecture that is constantly changing? We aim to address this question through an investigation of the effects of the short-term evolutionary process considering multiple weight initialization functions using CoDeepNEAT.

## 3 Methodology

We performed experiments with CoDeepNEAT on 4 weight initialization functions: (i) Glorot Normal; (ii) Glorot Uniform; (iii) He Normal; and (iv) He Uniform. To estimate reproducibility, we focused on well-known data sets with standard train/test partitioning, namely CIFAR-10 and MNIST. Our implementation is the same as provided in [8], with the addition of max pooling and batch normalization layers as possible components for the co-evolutionary algorithm.

**Table 1.** Hyper and evolutionary parameters. The Node Hyperparameters define the restrictions for CoDeepNEAT when creating new individuals.

| Node Hyperparameters | Range |
|---|---|
| Number of filters | [32, 256] |
| Dropout Rate | [0, 0.7] |
| Max Pooling | {True, False} |
| Batch Normalization | {True, False} |

| | | | |
|---|---|---|---|
| Generations | 5 | Training Epochs | 5 |
| Population Size | 10 | Blueprint Population Size | 10 |
| Module Population Size | 30 | Mutation Rate | 0.5 |
| Crossover Rate | 0.3 | Elitism Rate | 0.2 |

Each combination of initializers was run five times to reduce measurement bias. Because the data sets are fairly large and class-balanced, we evaluate the algorithms in terms of accuracy on their standard hold-out partitioning. The best individual in every step was fully trained for 100 epochs and then tested.

## 4 Results and discussion

For the CIFAR-10 dataset we perform the experiments using 5000 images for the evolution process (85/15 train-validation partition), while considering the fittest individual with the highest accuracy in the 10000 images from the standard test set. By the end of the genetic algorithm we fully trained the best individual and tested it in the test set again. Similarly, the steps were repeated for the MNIST dataset. The entire process was performed by a 4-GB GPU (GeForce GTX 1650/PCIe/SSE2), 15.5 GiB of RAM, in less than 42 sequential hours.

The statistical results show that in a short-term run the Glorot Uniform has a slight advantage among the other initializers for the CIFAR-10 dataset. For the MNIST dataset however, no advantage can be seen as all four initializers achieved results above 98% of accuracy. Even though the results are not among the start-of-art performances, we achieved above 99.20% in the MNIST and above 84.60% in the CIFAR-10 in less than a hour for both individual runs, with significantly less computational time, which suggests that focusing on short-term effects of the weight initialization functions is a promising direction for co-evolution of neural network architectures.

Table 2. Results obtained during experiments.

| CIFAR-10 | Run | Accuracy | MNIST | Run | Accuracy |
|---|---|---|---|---|---|
| Glorot Normal | 1 | 0.7681 | Glorot Normal | 1 | 0.9872 |
| Glorot Normal | 2 | 0.7861 | Glorot Normal | 2 | 0.9893 |
| Glorot Normal | 3 | **0.8057** | Glorot Normal | 3 | 0.9900 |
| Glorot Normal | 4 | 0.8034 | Glorot Normal | 4 | 0.9893 |
| Glorot Normal | 5 | 0.7995 | Glorot Normal | 5 | **0.9903** |
| Glorot Uniform | 1 | **0.8462** | Glorot Uniform | 1 | 0.9890 |
| Glorot Uniform | 2 | 0.7188 | Glorot Uniform | 2 | 0.9908 |
| Glorot Uniform | 3 | 0.8118 | Glorot Uniform | 3 | 0.9876 |
| Glorot Uniform | 4 | 0.7654 | Glorot Uniform | 4 | **0.9914** |
| Glorot Uniform | 5 | 0.8180 | Glorot Uniform | 5 | 0.9881 |
| He Normal | 1 | 0.7798 | He Normal | 1 | 0.9879 |
| He Normal | 2 | 0.6894 | He Normal | 2 | 0.9910 |
| He Normal | 3 | 0.7159 | He Normal | 3 | 0.9870 |
| He Normal | 4 | 0.7706 | He Normal | 4 | **0.9921** |
| He Normal | 5 | **0.7919** | He Normal | 5 | 0.9899 |
| He Uniform | 1 | 0.7743 | He Uniform | 1 | 0.9902 |
| He Uniform | 2 | 0.7767 | He Uniform | 2 | **0.9923** |
| He Uniform | 3 | 0.7324 | He Uniform | 3 | 0.9903 |
| He Uniform | 4 | 0.7695 | He Uniform | 4 | 0.9900 |
| He Uniform | 5 | **0.7998** | He Uniform | 5 | 0.9870 |

# Using a bio-inspired model to facilitate the ecosystem of data sharing in smart healthcare

Yao Yao[1,2*][0000-0002-5882-0058], Meghana Kshirsagar[1,2*][0000-0002-8182-2465], Gauri Vaidya[3][0000-0002-9699-522X], Conor Ryan[1,2][0000-0002-7002-5815]

[1] Biocomputing and Developmental Systems lab, University of Limerick, Ireland
[2] Lero, the Science Foundation Ireland Research Centre for Software, Ireland
[3] Principal Global Services, Pune, India

`Yao.Yao@ul.ie, Meghana.Kshirsagar@ul.ie`

**Abstract.** Following the development of Information Technology (IT) techniques, data and the knowledge behind data have been increased exponentially today. The goals to better manage and share such massive big data become more and more critical in many prominent Artificial Intelligence (AI)-based smart industries. Smart healthcare is a typical example in these cases. Efficient management of data sharing apparently can lead to better diagnosis, illness prevention and epidemic monitoring. However, appropriate and robust management of sharing sensitive data among different stakeholders operating within the ecosystem still poses challenges of geographical boundaries and compliance to diverse data sharing and access rules across continents. To address the complexity of large-scale data sharing and provide an efficient solution, our study proposes a bio-inspired autonomic agent-based framework capable of leveraging the large-scale distributed data sharing infrastructure with multiple stakeholders whilst, at the same time, supporting the development of future data integration ecosystems. Our current study selects smart healthcare as a use case to discuss the deployment of our data sharing framework and use this example to demonstrate the potential advantage of this framework but the framework itself also has good extensibility to other interdisciplinary scenarios and domains.

**Keywords:** Bio-inspired Agents, Evolutionary Computation, Data Sharing

## 1  Introduction

The exponentially increasing digital data in the recent decades has inspired the research community to revolutionize industrial networks [1]. Multiple data sources in diverse formats had led to the need of establishing data sharing techniques to integrate these sources with secure fashion [2]. The near future aims at integration of data across multiple industrial systems to form a hyperconnected global network. Integration of data across multiple components of a system can be applied in multiple contexts like websites and cloud storages, e-commerce platforms, financial institutions, healthcare systems, land records and so on. The current data sharing models in industries rely on the internal network of the systems to drive the business opportunities [3]. However, efficient, robust and secure data sharing techniques can unlock huge potentials of the industries in terms of boosting productivity, creating new business opportunities, new customer experiences, accelerating independent research analysis, strengthening collaborations, with minimal resources. Moreover, more and more intensive collaboration based in complex networks even requires the data to be shared across the international industrial boundaries to augment intelligent decisions with full transparency. All of these give us a great challenge and opportunity to implement the capable data sharing framework on a large scale with massive transactions.

One of the main difficulties to build such a framework is the complexity and customized context on large scale data sharing. Many stakeholders in the system can have various requests and standards to collaborate with. To specify all necessary routines and running the daily regulation on such a framework will cost a huge effort which may be infeasible to do in many cases. Here, we try to introduce a bio-inspired model to help people to tackle this problem and build a self-organizing data sharing framework with automatic agents. These agents are evolved by incorporating evolutionary computation concepts mimicking the food chains found in the natural environment. The concrete solution can be accomplished through design of agents for performing key functionalities such as integrating data from diverse sources, embedding personal history of individual patients and user authentication. Some agents can also embed a set of rules for controlling access to information being shared among stakeholders operating within the ecosystem.

This paper discusses the present undergoing work by the authors on how agents can be evolved to manage efficient data sharing within healthcare industries. The deployment of these agents in healthcare and any similar data sharing ecosystem can enable communities of providers by supporting functions that can improve the operations of the entire system.

For instance, in the given healthcare scenario, sharing patient data among healthcare providers can assure smart healthcare services, however, with a probability of health information being misused if leaked across the network. Developing solutions enabling the sharing and integration of data, while preserving security and consent of the users before sharing any data across the network is the need of the hour. This has led to wide use of Electronic Health Record (EHR) systems to integrate patient history, current medications, immunizations, laboratory results, current diagnosis, etc. to develop a centralized network among the healthcare providers. Data integration to facilitate interoperability among stakeholders of the healthcare ecosystem brings in potential advantages for delivering smart healthcare. The patient's medical history can be shared with the doctor in his next visit, the medicines prescribed can be shared with the pharmacist while the bills in the hospitals and labs can be verified and shared with the insurance providers. An agent-based model (ABM) [4] that has been adopted here is a bio-inspired mechanism to automate the systems with interactive processes. This model simulates the internal environments of the systems and produces intelligent decisions on the basis of logical reasoning. ABM systems have the potential to simulate real-world human decisions, behavior, social environments making them robust to be widely used across all domains. With such agents, the system can automate the processes in the context of data sharing within smart healthcare systems while establishing the consensus of data sharing by massive interactions.

## 2   Bio-inspired collective agents imitate the food chain model in nature

The sharing issue with multiple stakeholders is common in the biological world and the food chain model can be regarded as one of the most typical examples. In a healthy ecological system, the self-organizing food chain model can perfectly solve and regulate the energy sharing in an extremely complex network while keeping the flexible equilibrium in the network. According to the previous researches [5,6], the efficiency of the natural food chain model is related to its special construction manner. The model comes from the bottom-up collective interaction of various biological organisms and each kind of organism can be regarded as one kind of stakeholder in the ecosystem. Through the interaction between stakeholders, such a model is self-organized gradually and it can dynamically keep optimized for adapting to the current environment. For achieving similar efficiency and robustness in our data sharing framework, we adopted the same bottom-up construct principles in the bio-inspired agents and use such agents to regulate the communication between these distributed stakeholders in the framework. Each user in our data sharing framework will have their own customized agent which will acknowledge the corresponding user's role in the data sharing. Such knowledge includes the interests of the user, data sharing schema, certification list, the description of resources with the user and so on. The knowledge can be initially given by the user and then automatically evolving by agents. During the data sharing, the agent will interact with other agents based on their knowledge about the particular user to share the data. When a user manually initializes a new transaction or inputs new instructions, the agents are also able to update their own knowledge simultaneously. The updated change on user's attributes will be disseminated through the interaction between agents.

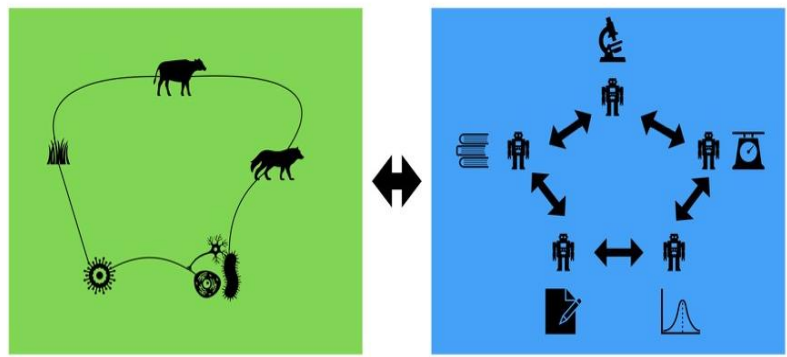

**Fig. 1.** Analogy of the food chain model and our bio-inspired data sharing framework

4## 3 Self-organizing data sharing framework and dynamic regulation in smart healthcare

Smart healthcare systems are in a position to use digital technology to improve patient outcomes while also improving operational efficiency. Smart Healthcare systems or infrastructures as depicted in Fig. 2 aim at delivering effective services by linking and coordinating the information among the individual components of the system as a whole. Though vast amounts of digital data generated across systems can augment the overall efficiency, they are at stake being leaked across the communication channels [7].

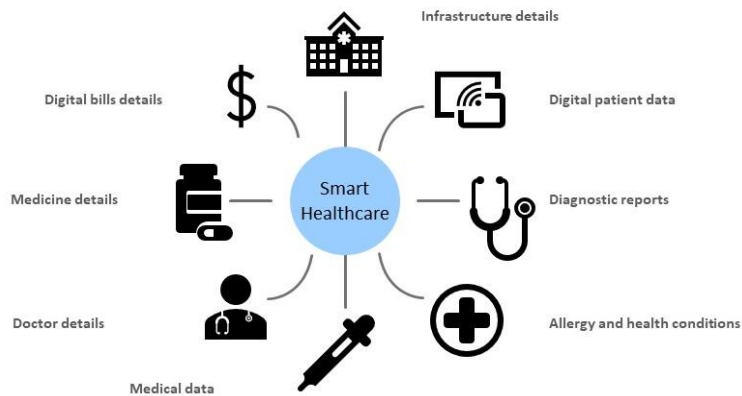

**Fig. 2.** The components of Smart Healthcare Systems

In our study, we use bio-inspired agents to regulate the data sharing transactions. The knowledge of every user's profile can be updated and regulated through the interaction between agents and knowledge bases. We have applied this framework on the data sharing of healthcare. With a blockchain network, the knowledge of regulating user's data sharing transactions will be well protected and its development is transparent and traceable to all users. The efficiency of the framework can also be improved by the evolvable agents and its particular knowledge. Fig. 3 below shows an example how the bio-inspired agents cooperate with each other in the data sharing of healthcare scenarios.

The proposed architecture is as shown in Fig. 3. There are vast amounts of patient data coming from two primary sources- the patient himself through use of mobile and wearable devices, medical devices implanted during surgical procedures, routine visits for general health checkup to clinics and medical practitioners. The other source is the clinical data associated with the patient such as scans, laboratory results, prescriptions for medications and other service providers belonging to healthcare industries. These data reside in various heterogeneous databases. Various autonomous agents are defined for key processes in healthcare such as: integration of data from diverse sources,

EVO* 2021 - Proceedings in ArXiv - Late-Breaking Abstracts  28trueheader

## 3 Self-organizing data sharing framework and dynamic regulation in smart healthcare

Smart healthcare systems are in a position to use digital technology to improve patient outcomes while also improving operational efficiency. Smart Healthcare systems or infrastructures as depicted in Fig. 2 aim at delivering effective services by linking and coordinating the information among the individual components of the system as a whole. Though vast amounts of digital data generated across systems can augment the overall efficiency, they are at stake being leaked across the communication channels [7].

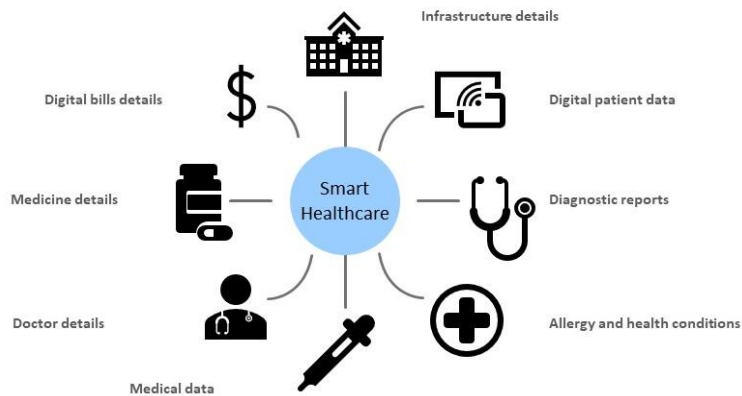

**Fig. 2.** The components of Smart Healthcare Systems

In our study, we use bio-inspired agents to regulate the data sharing transactions. The knowledge of every user's profile can be updated and regulated through the interaction between agents and knowledge bases. We have applied this framework on the data sharing of healthcare. With a blockchain network, the knowledge of regulating user's data sharing transactions will be well protected and its development is transparent and traceable to all users. The efficiency of the framework can also be improved by the evolvable agents and its particular knowledge. Fig. 3 below shows an example how the bio-inspired agents cooperate with each other in the data sharing of healthcare scenarios.

The proposed architecture is as shown in Fig. 3. There are vast amounts of patient data coming from two primary sources- the patient himself through use of mobile and wearable devices, medical devices implanted during surgical procedures, routine visits for general health checkup to clinics and medical practitioners. The other source is the clinical data associated with the patient such as scans, laboratory results, prescriptions for medications and other service providers belonging to healthcare industries. These data reside in various heterogeneous databases. Various autonomous agents are defined for key processes in healthcare such as: integration of data from diverse sources, authentication validation among network users, instantiating smart filters among legal entities within the network.

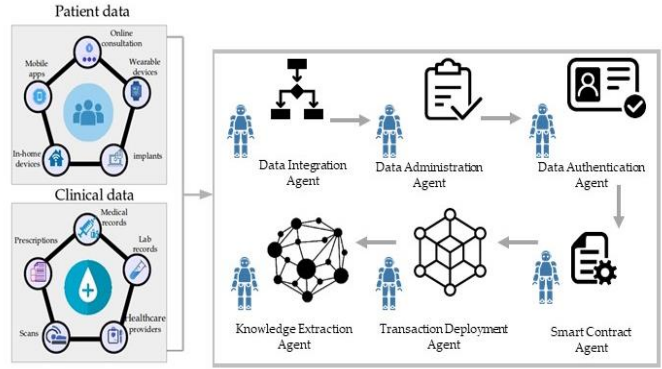

**Fig. 3.** Different agents within a smart healthcare system

Behind the agents, the system uses the knowledge of knowledge graphs to guide the behavior of each agent. Such knowledge includes the description of smart filters, the path of resources, the consensus of data administration and accessing. Agents can change and update the knowledge or implementing functions on the blockchain, meanwhile all agent's behaviors, changes and data will be recorded and protected in the blocks. As we have discussed above, the whole structure of data sharing framework is based on the interaction of all agents and it can be dynamically changed by altering knowledge of agents. Through an evolutionary algorithm embedded in each agent, the agent can evolve its own behavior model to better adapt to the current given environment for sharing data. All interaction between individual agents and users self-organizes the whole framework from a bottom-up manner.

## 4  Conclusions

This article described how to use computational bio-inspired agents to access the different distributed datasets and keep the data and knowledge sharing efficient, robust and compatible with all stakeholders in a smart healthcare system. This proposed framework provides a potential interdisciplinary solution for data sharing on many domains where it may need to implement intensive collaboration based on common data. Like the analogy of the natural food chain model, our proposed framework also inherits similar advantages to its biological counterpart. It is capable of dynamically adapting to the customized requests of various users and reaching a consensus through massive interaction. In the next step of this study, we will continually test the framework in a series of subsequent scenarios and initialize the interdisciplinary collaboration with the experts from different domains. Within the future collaboration, we will develop the corresponding benchmark tests based on the expert's suggestion to examine the functionality of the framework.

# Generating Music with Extreme Passages using GPT-2

Berker Banar[1,2] and Simon Colton[1,3]

[1] Game AI Group, EECS, Queen Mary University of London, UK
[2] Centre for Digital Music, EECS, Queen Mary University of London, UK
[3] SensiLab, Monash University, Australia

## 1 Introduction and Methodology

Music generation using machine learning is an active research field [1] [5] [6] and a typical approach is to use a symbolic music representation scheme, such as MIDI [3]. Recent advances in deep learning models for natural language processing have been employed in music generation projects [9] [8], e.g., MuseNet, which uses the GPT-2 generative text model [7]. Building on this work, we present a generative deep learning tool that can produce symbolic musical compositions with interesting passages that can be considered extreme in various ways, and a novel method for controlling the generation of symbolic music. We employ the GPT-2 version with 124m hyperparameters [8], which we fine-tune to various degrees with music data. Fine-tuned models are seeded with short musical excerpts to initiate the generation process. Using this to generate thousands of musical segments, we apply some musical analysis routines to categorise them, so that users can select segments in terms of how extreme they are, then combine them into final pieces.

We utilize four different models of GPT-2, each fine-tuned with the MIDI dataset from [4], which comprises 327 classical music pieces. A MIDI note is represented as text by mapping its note number (pitch), duration and start time (in that order) to string equivalents. Each neural model is fine-tuned to a different loss value, and some of the models are not trained well on purpose, so that they generate passages that would likely not be composed by a person. These passages can be considered as quite extreme, characterised as having one or more of: long and fast melodic sequences, big interval jumps and atypical rhythmic figures. Based on some initial experiments and our personal aesthetic considerations, we investigated a number of models with different loss values and settled on four of them. Taken together, the musical segments generated by these models cover a continuum from extreme to more mainstream passages.

Our tool uses the models to generate thousands of musical segments of around 5 seconds long. Then, it calculates average note duration, repetitiveness and musical interval jumps for each segment, which enables the curation and concatenation of a composition as a series of segments. To do this, we use sliders in a GUI to set the desired level for the metric and the system searches the pool for segments which satisfy these criteria. This has enabled us to compose pieces with particular arcs of extremity. In particular, we have been interested in pieces where the music builds to a crescendo of extremity towards the middle of the piece, releases this tension, and then becomes calmer towards the end.

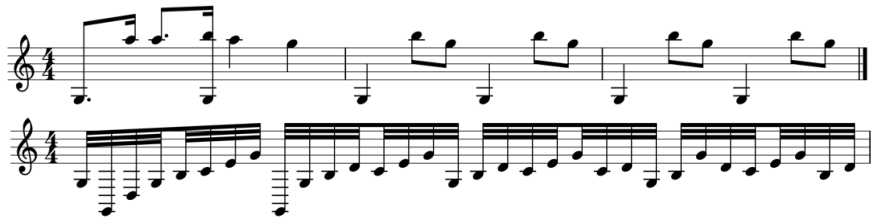

**Fig. 1.** Example extreme passages generated by the tool. In the top line, we see large melodic intervals and repeated figures. In the bottom line, we see a single bar that is comprised only of 32nd notes.

In a second, fully automated, approach, we use the music at the end of one generated segment to seed the production of new segments, which can bring some level of continuity to pieces composed of chained segments in this way. Naturally, this requires generation and concatenation processes happening in turn, so user-directed interactive search over pre-generated segments is not feasible. Hence, the tool composes entire pieces via iterative seeding, and we use the values for the metrics to help us find interesting pieces among hundreds generated.

In both approaches, we encourage the generation of passages of music outside the norm in terms of sequences of extremely fast, jumpy and/or repetitive notes. In other contexts, these could be seen as defective and discarded as they don't reflect the musical distribution in the data particularly well. However, importing them into compositions in a controlled way could be musically interesting and a source of inspiration for new aesthetic territories. Of course, some of these extreme passages may be challenging to perform even for professional musicians. So, compositions with these extreme passages may be better performed by robot performers, such as Shimon [2], or with people and machines performing different passages of the same piece. Figure 1 depicts example extreme passages, in terms of melodic intervals, repetitive figures and note durations.

## 2   Experimental Results

In the first approach described above, compositions are shaped by a user as either 5 or 10 segments of roughly five seconds duration. Pre-generated segments can be searched for by interacting with the UI, in terms of required average note duration, repetition and interval values, set with three groups of five sliders on the interface, each of which has 8 levels as portrayed in Figure 2. When prompted by the user, the tool constructs sample compositions by concatenating the segments retrieved, and users can browse the samples. For each of the generated compositions, audio, midi, piano roll and musical score options are available for the user to assess the piece with. Figure 2 presents a screenshot of the composition tool with a curated piece, where the sliders are set to small interval jumps, a small number of repetitive notes and long note durations throughout the composition. A video demonstration of using the tool is available at: https://www.youtube.com/watch?v=sSrCUmr3buI.

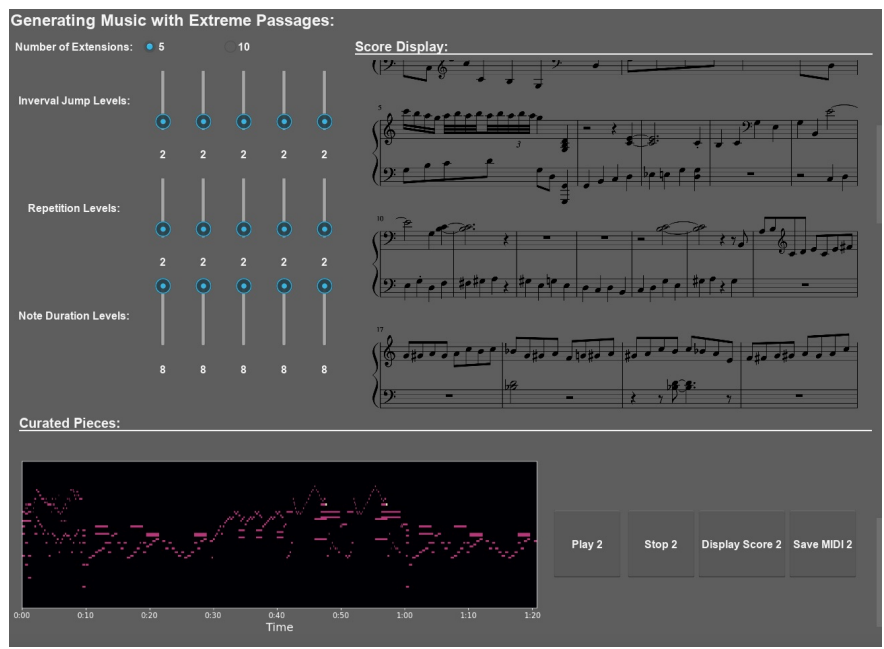

**Fig. 2.** A screenshot of the composition tool

Many traditional musical compositions have uncomfortable elements of discordant, atonal or rhythmically challenging music, the resolution of which adds to the pleasure and intrigue listeners have. In our context, we evaluated the generated content as often having interesting but extreme melodic and harmonic elements. Extreme passages can bring excitement, but having too many will likely be uncomfortable for most listeners. However, if they are introduced in appropriate amounts, then listening to such extreme passages can be a pleasing and even exhilarating experience. For our first compositions, we used the tool to create short pieces with an arc of extremity, reaching a maximum in the middle of the piece. Listeners know pieces are short, so the apex dissipates quickly, and any discomfort will be short-lived and resolved, hopefully in a pleasurable way.

In an interactive session with the tool, we used as the seed an excerpt from Haydn's Piano Sonata in C major, Hoboken XVI:7, 1st Movement. We generated 17,600 segments, with each of the four models contributing 4,400 segments. Using the tool, we composed a number of pieces with the extremity arc as above, making four of these are available here: (https://soundcloud.com/user-330551093/sets/generating-music-with-extreme). In the second, fully automated, approach above, we used the tool to generate 500 pieces, using the same initial seed, but with iterative re-seeding, as discussed. Six of these – which are reminiscent of the music of Philip Glass and sequential music in general – were chosen for the same SoundClound page as above.

## 3  Future Work

We intend to investigate how to replace the musical analysis and categorisation system with a discriminative neural network, to add extra sophistication and automation to the process. We are also considering implementing a co-operative performance system so that human and AI players can play in a hand over (virtual) hand fashion, with the AI system handling the extreme passages. Moreover, as previously mentioned, we plan to use the composition tool to produce pieces specifically for robot performers such as Shimon [2], as they should be able to cope well with the unusual music. Also, we would like to explore the idea of feeding the post-processed outcomes back into the corpus to fine tune GPT-2, in the hope of designing new, interesting and surprising musical forms and styles. Ultimately, our aim is to investigate and promote the idea that novel computer-generated music of various sorts, that might have previously been discarded for being unlike human compositions, can have value in musical culture.

## Acknowledgements

Berker Banar is a PhD student at the UKRI Centre for Doctoral Training in Artificial Intelligence and Music, funded by UKRI grant number EP/S022694/1 and Queen Mary University of London.

## Supplementary Material

### Screenshots of the Composition Tool

In addition to Figure 2, there are two screenshots below, which have more extreme compositions than the one in Figure 2 due to the corresponding slider settings. The most extreme one is at the bottom here.

# Beauty: A Machine-Microbial Artwork *

Carlos Castellanos[1], Johnny DiBlasi[2], and Philippe Pasquier[3]

[1] Rochester Institute of Technology, Rochester, NY, USA
`carlos.castellanos@rit.edu`
[2] Iowa State university, Ames, IA, USA
`jdiblasi@iastate.edu`
[3] Simon Fraser University, Vancouver, BC, Canada
`pasquier@sfu.ca`

**Abstract.** We discuss *Beauty*, a hybrid biological-technological artwork, currently in development by Phylum, an experimental research collective an experimental research collective specializing in cultural production informed by the intersections of science, technology and the arts (and of which the authors are members of). The work is based upon an artificial intelligence agent that uses deep reinforcement learning to interact with alter cultures of pattern-forming social bacteria in order to make them more aesthetically pleasing.

**Keywords:** pattern-forming bacteria · deep reinforcement learning · computational aesthetic evaluation.

## 1 Introduction

We describe *Beauty*, a hybrid biological-technological artwork, currently in development by Phylum, an experimental research collective specializing in cultural production informed by the intersections of science, technology and the arts (and of which two of the authors are members). The work creates a situation where the fates of some contaminated soil and a group of bacterial cultures are determined by the whims of an artificial intelligence agent, which has an internal model of "beauty". The agent builds its model by observing the cooperative pattern-forming and swarming behaviours of selected bacterial species. It then attempts to modify the bacteria growth, both visually and spatially by introducing chemical attractants and repellents. As shown in Figure 1, cultures are placed under microscopes for observation and analysis by the agent. Images of their growth, movement and newly acquired synthetic abilities are captured using time-lapse microscopy. Growth patterns, colours and spatial dynamics are analysed by the agent to determine how well the colonies conform to its internal model of beauty. The more beautiful the growth patterns of the cultured bacteria appear to the agent the more of a remediating solution the soil receives and the more nutrients the bacterial cultures receive. It is known however, that these bacteria only produce their intricate patterns under environmental stressors such as lack of food and moisture. Thus, the agent also has to reduce nutrient levels and introduce stress-inducing chemicals (e.g. non-lethal concentrations of antibiotics) into the bacterial cultures. In addition, the bacteria will be genetically modified in a way that manipulates their stress-response genes to express the aforementioned "beauty enhancements". Thus in order to properly remediate the contaminated soil, the bacteria may have to starve themselves in order to look beautiful for the agent. The agent also expresses its "feelings" about this process via a series of evolving sound and visual patterns.

---

* Supported in part by Coalesce Center for Biological Arts, State University of New York at Buffalo, Rochester Institute of Technology and Iowa State University

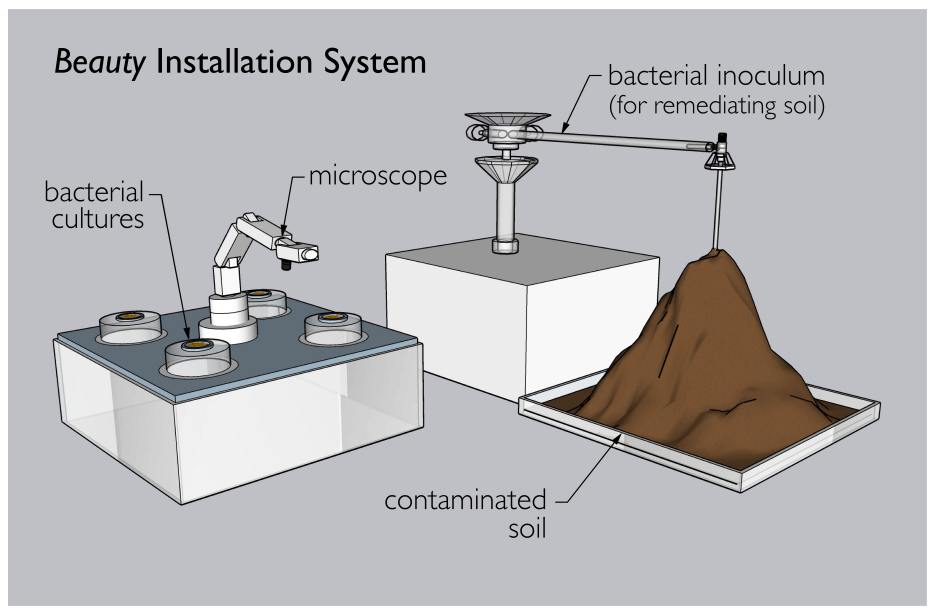

**Fig. 1.** Fig 1. *Beauty*, 2020, Phylum, microscope, robot arm, soil, Copyright by Phylum.

## 2   A Primer on Pattern-Forming Bacteria

Modern bacteriology abounds with examples of dynamic self-assembly and self-organization [7]. The focus of this project is on cooperative pattern-forming of bacterial colonies. These are characterized by the formation complex, often fractal-like, patterns that colonies develop in response to adverse growth conditions, such as lack of nutrients [2]. These occur most often in *Bacillus subtilis* and certain members of the genus *Paenibacillus* (such as *P. dendritiformis*, shown in Figure 2). While the aesthetic value of these patterns have not been studied in any significant way, they have nevertheless proven to be a source of curiosity and fascination for scientists and laypersons alike.

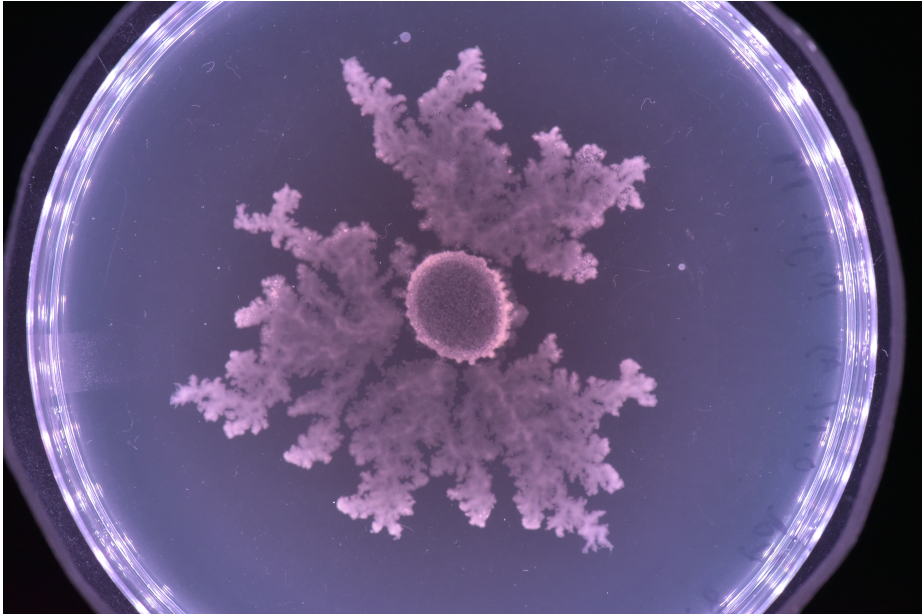

**Fig. 2.** Fig 2. A colony of Paenibacillus dendritiformis under dark-field illumination after approximately 48 hours of growth.

## 3 Agent Model

Our agent model is based upon the world model [5], a deep reinforcement learning model that generates simulated environmental states for the agent to adjust its policy (mappings of states to actions) before taking action in the real environment. A distinguishing feature of the world model when compared to other model-based RL approaches [6] is the ability to train an agent within its own internally generated model of the environment rather than the environment itself. In essence, the agent generates images of the environment and trains itself through experimentation within its own generative hallucinated dreams, rather than the actual environment. In our project, we are currently experimenting with extending the world model by adding a physics-based domain knowledge of the bacterial growth and an aesthetic evaluation model. This facilitates agent training and allows the agent to more reliably manipulate the bacterial colonies. In essence, enabling the agent to dream of other, more beautiful bacterial colonies and attempt to make those a reality. While we are currently testing our system with the "stock" world model, future iteration will include a physic-guided RNN [8] that uses the "communicating walkers" and lubricating bacteria models of bacterial growth and pattern-forming [3].

In addition, our research also involves the challenging task of developing computational aesthetic evaluation methods for the bacterial patterns. Computational aesthetic evaluation involves computers making normative judgements related to questions of beauty and taste in the arts [4]. This is a largely unsolved problem, with no agreed upon methods and metrics. We are currently exploring both the use of fractal dimensions – which have been shown to have some correlation to aesthetic preferences for chaotic patterns [1] – and conditioning our model on data sets of artworks that are considered "beautiful" (for example all of the paintings in the Louvre in Paris, or of people or things that are considered attractive or aesthetically pleasing in particular cultures).

## 4   Conclusions

This project addresses timely and relevant issues by establishing a unique interplay between "primitive" microorganisms, cultural notions of beauty and aesthetic judgement, the status and implications of intelligent machines and the impact of humans (and their technologies) on our ecology. In addition, while today all manner of microorganic labor is marshaled to produce products for humanity ranging from food to fuel to pharmaceuticals (not mention their use in cleaning up our environmental messes), we recognize these creatures as lively and dynamic, with agency and lifeworlds of their own. Thus the motivation for this work lies in creating an interface or window through which these organisms can convey their complexity and otherness using a language that can be understood by humans and the intelligent systems they create. We hope that this "creative misuse" of synthetic biology and AI will inspire new ways of looking at the relationship between humans, technology and the more than human world.

# Distributed species-based genetic algorithm for reinforcement learning problems

Anirudh Seth[1,2][0000−0003−3762−2578], Alexandros Nikou[2][0000−0002−8696−1536], and Marios Daoutis[2][0000−0003−0401−6897]

[1] KTH Royal Institute of Technology, Brinellvägen 8, 114 28 Stockholm, Sweden
[2] Ericsson Research, Torshamnsgatan 23, 16440, Stockholm, Sweden
{anirudh.seth, alexandros.nikou, marios.daoutis}@ericsson.com

**Abstract.** Reinforcement Learning (RL) offers a promising solution when dealing with the general problem of optimal decision and control of agents that interact with uncertain environments. A major challenge of existing algorithms is the slow rate of convergence and long training times especially when dealing with high-dimensional state and action spaces. In our work, we leverage evolutionary computing as a competitive alternative to training deep neural networks for RL problems. We present a novel distributed algorithm based on efficient model encoding which enables the intuitive application of genetic operators. Another contribution is the application of crossover operator in two neural networks in the encoded space. Preliminary results demonstrate a considerable reduction of trainable parameters and memory requirements while maintaining comparable performance with DQN and A3C when evaluated on Atari games, resulting in an overall significant speedup.

**Keywords:** Neuro-evolution strategies · Model Encoding · Distributed Speciation · Reinforcement Learning · Genetic Algorithms

## 1 Introduction

Reinforcement Learning (RL) is an active area of research within artificial intelligence routinely applied in a wide range of domains, for example robotic manipulation [1], personalized recommendations [10] and recently in novel areas such as telecommunication [6]. Existing methods, e.g. Q-learning [4] and policy gradients [3], train deep neural networks by back-propagation of gradients. Such approaches usually entail expensive computations, often not trivially parallelizable, which result in hours or even days of training in order to obtain desirable results especially when solving complex problems with large state-spaces [3, 4].

Alternative approaches to solve RL problems include Augmented Random Search (ARS) [2], evolution strategies [7] and deep neuro-evolution [9]. The latter two approaches belong to a class of black-box optimisation techniques which are based on principles of evolutionary computation. The results highlight the robustness of these approaches to sparse/dense rewards, tolerance to arbitrarily long time horizons as a consequence of gradient free learning and significant speedup due to distributed training [7, 9].

In our work, we propose a novel distributed species-based genetic algorithm aimed at solving RL problems, which trains a deep neural network using the genetic operators mutation, selection and crossover. The algorithm relies on an

2     A. Seth et al.

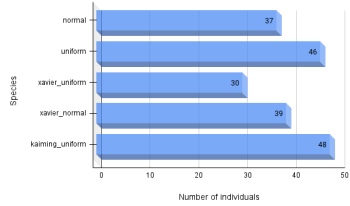
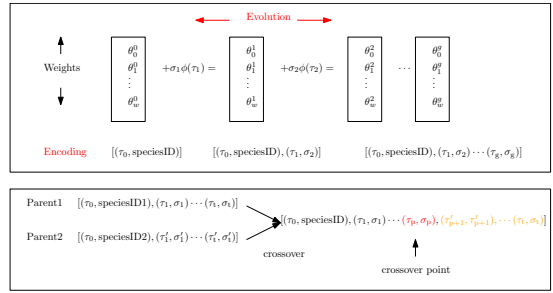

Fig. 1: Distribution of species in the initial population.

Fig. 2: Graphical representation of the model encoding, mutation and crossover.

efficient and compact model encoding that scales well in memory and utilizes extremely low bandwidth making it highly scalable. The performance is presently evaluated in Atari 2600 benchmarks and is compared to established algorithms such as DQN [5] and A3C [3].

## 2  Methods

As in any other typical genetic algorithm or population based optimisation method, we evolve a population $\mathcal{P}$ of $\mathcal{N}$ individuals (candidate solutions to the problems – i.e. a *DQN* network). The concept of speciation to cluster the individuals with topological similarities was first introduced in NEAT [8]. We build upon the same idea and randomly generate $\mathcal{S}$ classes of species, with each class represented by a unique NN layer weight initialisation strategy, that promotes diversity and protects innovation within the population pool. We call this algorithm *Sp-GA* in the following text.

At each generation, we evaluate the entire population using a fitness function (cumulative reward for a RL problem). A fraction of the population $\mathcal{T}$ with highest fitness scores are selected as parents for the current generation. These parents then generate the offsprings for the subsequent generation using genetic operators like mutations, crossover and selection. The mutation operator selects a single parent uniformly at random from $\mathcal{T}$ and updates all the parameters by applying additive Gaussian noise as $\theta' = \theta + \sigma\epsilon$ where $\epsilon \sim \mathcal{N}(0, I)$ and $\sigma$ is the mutation power. The individual with the highest fitness is copied as-is to the next generation, a concept called elitism. Crossover operation selects a pair from $\mathcal{T}$ to generate a new individual with features from each parent. The parameter $\psi$ represents the probability of using mutation and $1 - \psi$ for crossover.

In our work, we propose a novel way to perform single point crossover of two neural networks to produce one child neural network in the encoded space. This evolutionary process is repeated until a new population of size $\mathcal{P} + 1$(elite) is generated completing one epoch/generation of the genetic algorithm.

### 2.1  Model encoding and genetic operators

Authors in [7, 9] employ lists of random seeds for distributed computing but these approaches are limited to a vanilla genetic algorithm without crossover. In our work, we propose an enhanced model encoding that scales well in memory, can be easily serialized, utilizes low bandwidth and can also work with a wide variety of genetic algorithms. An individual from the population (a NN) can be represented as a list of tuples (Fig. 2), the first tuple contains the specie identifier

| Hyperparameter | Value |
|---|---|
| Population | 200 |
| Mutation Power | 0.002 |
| No of elites | 10% |
| $\Psi$ | 0.75 |
| Species | 5 |

|  | DQN | A3C | Sp-GA |
|---|---|---|---|
| No. of workers | 1 CPU | 8 workers - 1 CPU each | 10 workers - 4 CPU each |
| Frames | 5M | 5M | 5M |
| Training time | $\sim$ 9hr | $\sim$ 3hr45min | $\sim$ 55min |
| frostbite | 240 | 180 | **270** |
| spaceinvader | 585 | 555 | **1020** |

Fig. 3: Hyperparameters used to train Sp-GA

Fig. 4: Maximum episodic reward for DQN, A3C and elite's performance for Sp-GA after training on $\sim$ 5 million frames on 2 Atari 2600 games.

(class name), initialisation seed and the subsequent tuples represent the mutation power, seeds utilized for evolutionary operations at each generation. Given this encoded representation, a worker can decode the model by iterating through this list of seeds and applying iterative mutations.

Crossover of two neural networks is a challenging task to implement in a distributed manner. Previous work [7, 9] only rely on mutations when evolving the neural network. The chromosomal crossover and recombination of genes in cells motivated us to apply the same approach to the encoded representation of the networks. A crossover point is sampled and a new encoding is formed by merging the two encodings along the sampled point. The resulting encoding represents a valid individual which contains features from each parent (Fig. 2).

## 3 Results

The performance of the proposed algorithm (*Sp-GA*) was evaluated on Atari 2600 games. Due to limited computational resources, we compare the results from 2 games, i.e. *space-invaders* and *frostbite* with reference implementations [3] of *DQN* and *A3C*. The tuned hyper-parameters were obtained from the benchmarked results [4]. The following steps in our algorithm are identical to *DQN* and *A3C*: 1) data pre-processing 2) network architecture 3) 30 random, initial no-op operations The hyper-parameters used for training are summarized in Fig. 4 and the initial distribution and kinds of species are shown in Fig. 1. Each episode (*DQN* and *A3C*), generation *Sp-GA*) was capped at maximum of 10k frames. The maximum episodic reward (*DQN* and *A3C*) and elite individual's reward (*Sp-GA*) after training for $\sim$ 5 million frames on a Kubernetes cluster is reported in Fig. 4. The elite model for frostbite belongs to the species 'xavier_normal' and for space-invader it belongs to the species 'kaiming_uniform'. *Sp-GA* outperforms *DQN* and *A3C* on both games and is at least 4 times fast. However, all these comparisons are preliminary and need further investigation. The key takeaway is that *Sp-GA* is a comparable competitor to *DQN* and *A3C* and is significantly scalable.

## 4 Discussion

In this work we presented a distributed species-based genetic algorithm as a scalable alternative to RL algorithms like *DQN* and *A3C*. A novel framework to apply crossover in the encoding space is also proposed. Our results show comparable performance and significant speedup when compared to Markov Decision Process (*MDP*) based approaches. The results in our work are limited to 2 games and training on $\sim$ 5 million frames and require further verification. We have also demonstrated how different variants of genetic algorithms can be scaled and quickly solve RL problems with a large state space. Future efforts will be devoted towards training our algorithm on a wider set of games for more training frames and tuning the hyper-parameters for each use case. It is also interesting to see how this algorithm would perform in RL problems from other domains such as from robotics and telecommunication. Overall it is noteworthy to see a simple algorithm performing surprisingly well paving the way to novel variants such as a hybrid approach combining *Sp-GA* with a *MDP*-based algorithm.

---
[3] https://docs.ray.io/en/master/rllib.html
[4] https://github.com/ray-project/rl-experiments

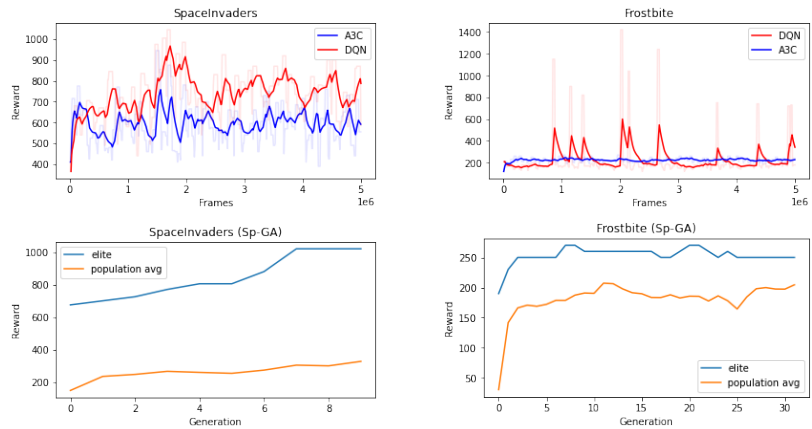

Fig. 5: Performance of DQN, A3C (max score over an episode), Sp-GA(top elite's score) with increasing epochs on Atari 2600 games.